\documentclass[letterpaper, 10 pt, conference]{ieeeconf}  

\IEEEoverridecommandlockouts                              

\usepackage{amsmath}
\DeclareMathOperator*{\argmin}{arg\,min}
\usepackage{diagbox}
\usepackage{pifont}
\usepackage{multirow}
\usepackage{mathtools}

\usepackage{floatrow}
\usepackage{graphicx}
\usepackage{caption}
\usepackage{subcaption}
\usepackage{hyperref}

\title{\LARGE \bf
Tracking monocular camera pose and deformation for \\ SLAM inside the human body
}

\author{Juan J. Gómez Rodríguez,
        J.M.M. Montiel,~\IEEEmembership{Member,~IEEE} and Juan D. Tardós,~\IEEEmembership{Fellow,~IEEE}
\thanks{This work was supported by EU-H2020 grant 863146: ENDOMAPPER, Spanish government grant PGC2018-096367-B-I00 and by Aragón government grant DGA\_T45-17R and PhD scholarship of J. J. Gómez-Rodrígez.}
\thanks{The authors are with the Instituto de Investigaci\'on en Ingenier\'ia de Arag\'on (I3A), Universidad de Zaragoza, 
Mar\'ia de Luna 1, 50018 Zaragoza, Spain. E-mail: \{jjgomez, josemari, tardos\}@unizar.es.}
}

\begin{document}

\maketitle
\thispagestyle{empty}
\pagestyle{empty}

\begin{abstract}
Monocular SLAM in deformable scenes will open the way to multiple medical applications like computer-assisted navigation in endoscopy, automatic drug delivery or autonomous robotic surgery. In this paper we propose a novel method to simultaneously track the camera pose and the 3D scene deformation, without any  assumption about environment topology or shape. The method uses an illumination-invariant photometric method to track image features and estimates camera motion and deformation combining reprojection error with spatial and temporal regularization of deformations. Our results in simulated colonoscopies show the method's accuracy and robustness in complex scenes under increasing levels of deformation. Our qualitative results in human colonoscopies from Endomapper dataset show that the method is able to successfully cope with the challenges of real endoscopies: deformations, low texture and strong illumination changes. We also compare with previous tracking methods in simpler scenarios from Hamlyn dataset where we obtain competitive performance, without needing any topological assumption.

\end{abstract}

\section{INTRODUCTION}

Visual Simultaneous Localization and Mapping (SLAM) and Visual Odometry (VO) in static environments have been hot research topics in the last decades and many methods have raised to solve them with outstanding accuracy and robustness using features \cite{mur2015orb}, direct methods \cite{engel2017direct}, or hybrid techniques \cite{forster2016svo}. The increasing popularity of these techniques has raised expectations to solve SLAM in more complex scenarios. For example, one can think of many useful applications of SLAM in Minimal Invasive Surgery (MIS) like guiding surgeons through augmented reality annotations towards the place where a polyp was detected in a previous exploration, and automatic polyp measurement to analyze its evolution. Moreover, surgical robots would greatly benefit from SLAM inside the human body as they will be more secure, robust and accurate, and they will be able to combine information coming from previous explorations or from other sensors like Computerized Tomography (CT).

\begin{figure}
    \centering
    \includegraphics[width=0.95\linewidth]{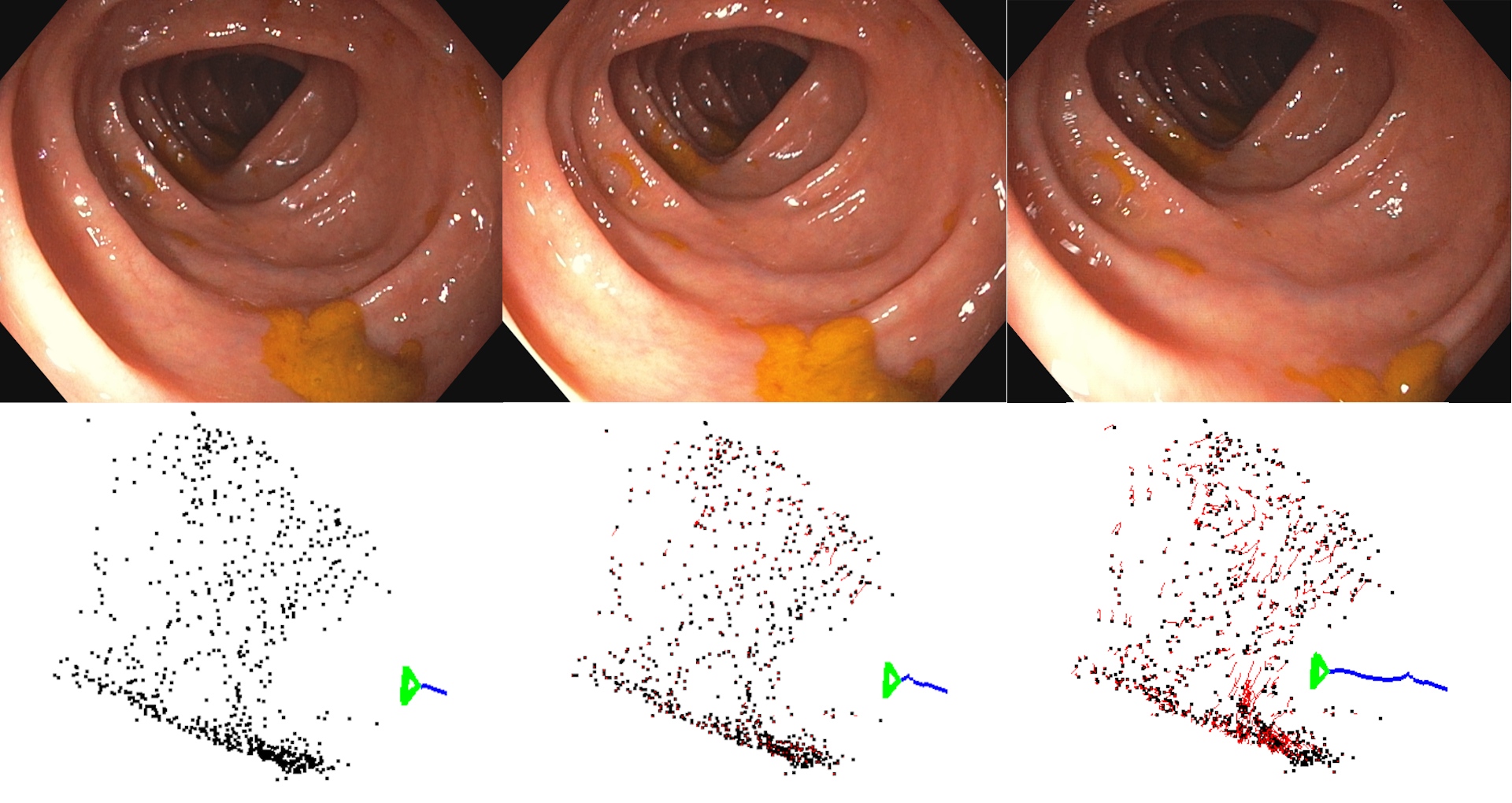}
    \caption{Tracking the camera pose and scene deformation in a human colonoscopy from Endomapper dataset. Top: images from the sequence. Bottom: camera trajectory in blue, undeformed map points in black and map point deformation trajectories in red.}\label{fig:Tracking_Endomapper}
\end{figure}

However, visual SLAM inside the human body poses tremendous challenges like weak texture, changing illumination, specular reflections, and lack of rigidity (Fig. \ref{fig:Tracking_Endomapper}). 
Weak texture and specular reflections hinder data association algorithms based on feature matching, preventing methods like ORB-SLAM3 \cite{campos2021orb} from working in these sequences. On the other hand, changing illumination puts direct methods like DSO \cite{engel2017direct} or DSM \cite{zubizarreta2020direct} and hybrid methods like SVO \cite{forster2016svo} in serious trouble, as they assume constant illumination of the environment. In contrast, we solve data association with a modified Lucas-Kanade algorithm, first presented in \cite{gomez2021sd}, that is able to cope with local illumination changes.

But the major challenge to be addressed for SLAM inside the human body is deformable scenes, as breaking up with the rigidity assumption impairs both environment reconstruction and camera tracking.
The recent DefSLAM \cite{lamarca2020defslam} is the first monocular deformable SLAM system able to perform tracking and mapping, but it strongly relies on the assumption of a smooth continuous shape with planar topology, which does not hold in colonoscopies (see Fig. \ref{fig:Tracking_Endomapper}).



In this paper we present the first pure monocular method able to initialize a map and track camera pose and scene deformation in general scenes inside the human body (Fig. \ref{fig:Tracking_Endomapper}), without any topological or shape assumption. Our main contribution is a simple formulation that combines photometric feature tracking and an optimization based on reprojection error with spatial and temporal regularizers that encode local assumptions over the environment deformation, endowing our algorithm with enough expressivity to model complex scenes and track their deformations in real-life endoscopies. We provide quantitative evaluation on realistic colonoscopy simulations \cite{incetan2021vr} and qualitative results on real human colonoscopies from the Endomapper project \cite{Endomapper}, that were out of reach for previous techniques. We present quantitative comparisons in almost-planar scenes from Hamlyn dataset \cite{mountney2010three} where we obtain competitive performance, despite not using any assumption on the scene topology or shape.


\section{RELATED WORK}\label{sec::related_work}
The computer vision and robotics communities have developed excellent rigid visual SLAM systems \cite{engel2017direct} \cite{forster2016svo} \cite{campos2021orb} in the last years. While all these algorithms use quite different techniques they all rely in a vital, yet simple, assumption: that the environment is static. In contrast, deformable SLAM, which completely breaks up with the rigidity assumption, is still a challenging research topic.

Many works have tried to solve deformable SLAM by using sensors that provide complete 3D information of the environment like stereo or RGB-D cameras. This is the case of the seminal DynamicFusion \cite{newcombe2015dynamicfusion} which uses RGB-D images to reconstruct highly deforming environments with an Iterative Closest Point (ICP) algorithm and a spatial regularizer to constrain deformations of points that are close to each other, that we adopt in our work. Several extensions to DynamicFusion have been developed since then, being the most notably VolumeDeform \cite{innmann2016volumedeform} which combines the use of SIFT features and reprojection error with a dense ICP data term, to reduce drift and increase robustness. In \cite{slavcheva2018sobolevfusion} they formulate a variational method that takes RGB-D images to reconstruct a deforming environment. This work is later extended in \cite{slavcheva2020variational} introducing camera pose computation. 

There is increasing interest in SLAM in Minimally Invasive Surgery (MIS), where RGB-D sensors are not available. For this reason, works like \cite{song2017dynamic} \cite{zhou2021emdq} use depth coming form stereo images and an error function that combines reprojection errors and regularizers to perform deformable SLAM. As before, the reprojection error is augmented using other 3D terms like ICP errors or Point-to-Plane errors. Regarding the regularizers used, they are similar to the one introduced in DynamicFusion to represent that deformations occur locally, using pair-wise deformation terms between close points, which prevents divergence of individual points in the reconstruction.

Nevertheless stereo cameras are not appropriate for certain applications like colonoscopies where two cameras with enough baseline may not fit in the body cavities. In this kind of scenarios, only monocular deformable SLAM can be performed. This is even a harder problem since no real 3D information is available from a single view, scale is unobservable, and combining multiple views of a deforming scene is an open issue. The first monocular deformable SLAM system is DefSLAM \cite{lamarca2020defslam} which splits the deformable SLAM problem in 2 threads, one for tracking and one for mapping. They use ORB features and  minimize a reprojection error term with a deformation energy term that penalizes stretching and bending of the imaged surface. However, as ORB features are quite unstable in intracorporeal sequences, SD-DefSLAM \cite{gomez2021sd} extends DefSLAM to a semi-direct method by integrating an illumination-invariant Lucas-Kanade tracker to perform data association, achieving better robustness and accuracy. Crucially, both methods assume that the surface has planar topology, and model the surface with a triangle mesh which impose a strong global condition over the environment: the imaged surfaces have to be continuous with no holes. This is quite a strong assumption that seriously limits the kind of scenes that can be handled by both algorithms excluding, for example, colonoscopies (Fig. \ref{fig:Tracking_Endomapper}).

To tackle this limitation, \cite{lamarca2021direct} proposes a fully photometric algorithm to track camera pose and deformation using sparse 3D surfels (surface elements) under the assumption of local isometry. The use of surfels that have no constrains between them allows to model any kind of topologies. While obtaining very good results in medical scenes, the method still requires 3D information coming from a stereo camera to initialize the surfels. Also, the use of large surfels (in practice, square patches of $23\times23$ pixels in the image) can easily violate the local isometry assumption and is inefficient as using too many close pixels provide redundant information  with little to no improvement in accuracy \cite{engel2017direct}. Furthermore, the regularizers used impose small deformations with respect to a pose at rest, which can be inappropriate in many applications. 

In contrast, in this work we propose a pure monocular method for tracking camera pose and deformation. Under the assumption of slow deformations, we perform fully automatic monocular map initialization to obtain a first seed of the environment structure. Following previous works \cite{forster2016svo,gomez2021sd}, we use photometric feature tracking for robustness and accuracy, and reprojection error  for convergence and efficiency during optimization. In addition, we integrate two regularizers that encode our assumptions of smooth and slow deformation in order to constrain the reconstruction problem. 


\section{DEFORMABLE TRACKING}\label{sec::DefTrack}
This section is devoted to presenting our tracking algorithm. We first present the assumptions that governing our system. Afterwards we introduce our data association for tracking. Then we explore our algorithm for monocular map initialization and finally we present our optimization backbone and its formulation encoding each one of our assumptions.

\subsection{Assumptions}
The biggest difficulty when dealing with deformable scenarios is that the rigid assumption is violated. This makes camera pose and deformation prediction a non-separable problem for which infinite solutions arise, i.e. not all degrees of freedom (DoF) are observable. 

This is even drastically worse when using a single monocular camera as the scale is also unknown. For all this, one must incorporate some \textit{a-priori} knowledge into the problem in order to confine the possible solutions into a reduced set of solutions that correctly represents the real nature of the environment. In this paper, we propose the following assumptions to constrain our reconstruction problem:

\begin{enumerate}
    \item\textbf{Local isometry}: we assume that the vicinity of a surface point follows an isometric model, that is local distances are preserved.
    \item \textbf{Smooth deformation}: 
    we consider that points that are close in space must undergo similar deformations.
    \item \textbf{Slow deformation}: deformations are assumed to happen slowly over time. 
    \item \textbf{Camera motion is faster than deformation}: finally, we assume that camera can move faster than deformations, so we attribute rigid motions to the camera, computing deformations as small as possible. 
    
\end{enumerate}

Assumption (1) enables us to use a phtometric feature tracker defining a small neighbourhood around each tracked point that is assumed to be locally rigid. This allows us to take an approach similar to \cite{lucas1981iterative} to perform short term data association. Assumption (2) introduces local constrains in the deformations observed without imposing a global deformation or surface model. This effectively makes our system general enough to model any environment. Assumption (3) allow us to imposes temporal continuity in the position of surface points, reducing the effect of image and data association noise. Finally, Assumption (4) is the one that allows us to separate camera motion and environment deformation. In all SLAM systems, the sensor provides relative information, and as a result, the absolute pose of camera and environment is not observable. In rigid SLAM this is simply addressed by choosing an arbitrary global pose, for example the first camera pose is chosen to be zero. In deformable SLAM this is not enough as a camera motion is 
indistinguishable from a hypothetical case where all the environment moves rigidly, what is called the {\em floating map ambiguity} in \cite{lamarca2021direct}. This assumption allow us to use regularizers that penalize deformation over camera motion. In that way, the rigid part of the relative motion between environment and camera will be attributed to camera motion, obtaining deformations as small as possible.

It is important to note that none of the above assumptions impose global constrains over the surface topology, smoothness, or deformation, allowing us to model generic deformations and environments.

\subsection{Data Association}
Our previous experience in monocular deformable SLAM \cite{gomez2021sd} has proven that an accurate data association is crucial in order to reach good accuracy and robustness. Indeed other works have shown the potential of direct methods in this task like \cite{engel2017direct} in which the photometric term allows to get as a byproduct of the tracking the feature associations. This is done by imposing a global rigid transformation to all points as it is assumed that the environment is stationary. However this is far to be true in deformable SLAM. Indeed one can not impose any global constrain to the data association step as it is easily violated by deformations.

\begin{figure}
    \vspace{2mm}
    \centering
    \includegraphics[width=0.90\linewidth]{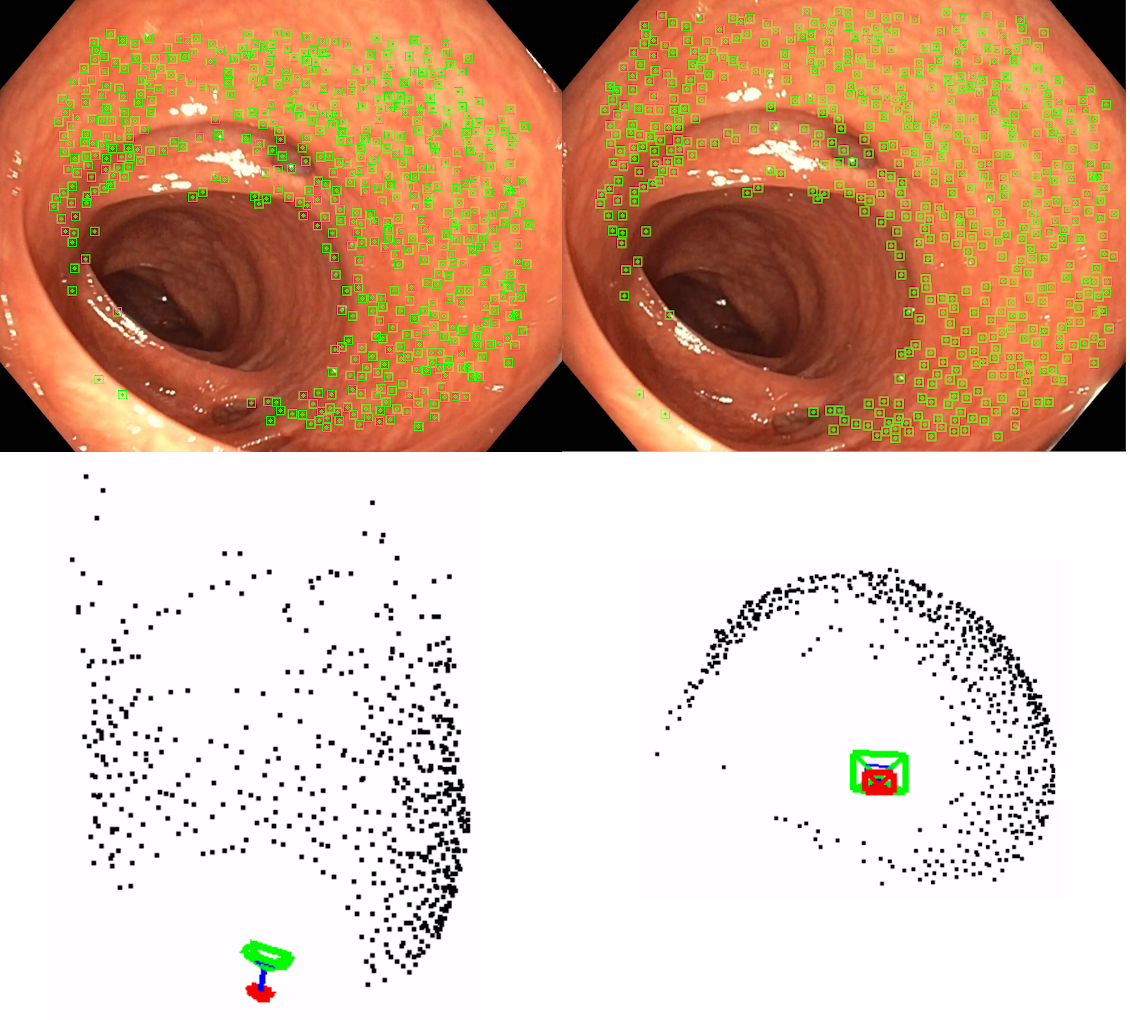}
    \caption{Top: Two images separated by 3 frames from our EndoMapper dataset with tracked features. Bottom: map initialized from those tracks}
    \label{fig::init}
\end{figure}

For that, we propose to perform photometric data association with Shi-Tomasi features \cite{shi1994good} prior the camera pose and deformation estimation using the modified multi-scale Lucas-Kanade algorithm proposed in \cite{gomez2021sd}:

\begin{equation}\label{eq::klt_enhan_residual}
\argmin_{\mathbf{d},\alpha,\beta} \sum_{\mathbf{v} \in P(\mathbf{u})}(I^{0}(\mathbf{v}) - \alpha I^t\left(\mathbf{v} + \mathbf{d}) - \beta\right)^2 
\end{equation}

where $P(\mathbf{u})$ is a small pixel patch centered at the keypoint $\mathbf{u}$. $I^0$ is the firs frame, where the points are intialize, and $I^t$ is the current frame in time $t$. These patches are updated every 5 images to account for big scale changes or rotations. This algorithm has been proven to achieve excellent results when tracking image features in short time steps even in the presence of deformations or local illumination changes (Fig. \ref{fig::init}). The key of its performance lies in using no global model: each point can move freely with respect the others. Also a local illumination invariance is achieved by computing local gain $\alpha$ and bias $\beta$ terms for each point.

In order to remove any possible outlier track, we compute the \textit{Structural Similarity Index} (SSIM) \cite{wang2004image} between the reference $x$ and tracked $y$ pixel patches to identify any outlier track:

\begin{equation}\label{eq::SSIM}
    SSIM(x,y) = \frac{(2\mu_x\mu_y+C_1)(2\sigma_{xy}+C_2)}{(\mu^2_x + \mu^2_y + C_1)(\sigma^2_x+\sigma^2_y+C_2)}
\end{equation}

where $\mu_x$ and $\sigma_x$ are the mean and covariance of the pixel patch, $\sigma_{xy}$ is the crossed covariance between both patches and $C_1$ and $C_2$ are constant values to avoid inestability when means and covariances approaches to zero. This has been proven to be a good similarity metric for small pixel windows as it combines in a same metric a luminance, contrast and structure comparison.

\subsection{Monocular map initialization}
\label{sec:monocular:initialization}


Initializing a map from monocular images in rigid environments is well known in Structure from Motion (SfM).  In deforming environments, Non-Rigid Structure from Motion (NRSfM) techniques can be used \cite{lamarca2020defslam}. However, they require, for example in the map initialization, assumptions such as a smooth scene surface with planar topology, which are not met in real colonoscopies (see Fig. \ref{fig:Tracking_Endomapper} and \ref{fig::init}). 

We propose to exploit assumption (4) using two close frames in which the environment can be considered quasi-rigid, and most image innovation can be attributed to the camera motion. This allows to apply SfM to obtain a first estimation of the map as if it is rigid and treat any deformation as small noise. 

Ideally, the method should to be independent of the camera model either pinhole or fish-eye. We propose to initialize the monocular map by computing the  Essential matrix between 2 close frames using as input normalized projective rays from features in the images. Our proposed initialization algorithm goes through the following steps:

\begin{enumerate}
\item Extract Shi-Tomasi features evenly distributed in the reference frame $I^{0}$ and track them in the current frame $I^t$ using the Lucas-Kanade optical flow algorithm. Unproject the matched features into normalized rays $\mathbf{x}_i^0$ and ${\mathbf{x}^t_i}$ and using the camera model unprojection function.
\item Compute an Essential matrix that relates poses of the two frames: \begin{equation}
{{\mathbf{x}}^t_i}^T \mathbf{E} \mathbf{x}_i^0 = 0
\end{equation}
This is done inside a RANSAC scheme to reduce the influence of outliers coming from the data association.
\item Recover the relative camera motion $\mathbf{T}_{C^tC^{0}}$ from $\mathbf{E}$. This will yield to 4 motion hypothesis (2 rotations and 2 translations). We are using close frames to initialize, hence the camera rotation should be small so we can safely select the smallest rotation to solve the rotation ambiguity. Finally we disambiguate the translation component by selecting the one that yields to the highest number of points in front of both cameras.
\item Reconstruct environment using the camera motion recovered. For that, we use the Inverse Depth Weighted Midpoint \cite{lee2019triangulation} to triangulate tracked features as it provides low 3D-2D errors in low parallax scenarios.
\end{enumerate}

\subsection{Camera pose prediction}
To encode assumption (4), we first estimate a preliminar camera pose $\mathbf{T}_{C^tW}$ for time $t$ prior estimating any deformation. We assume that the camera follows a physical model of constant velocity. This provides us with an initial guess of the camera pose that will then be refined using Non Linear Least Squares (NLLS) using a reprojection error with the environment geometry observed in the previous frame $t-1$. This can be seen as a way to attribute to a camera motion most of the image innovation seen. Note that this is not the final pose we compute but just a seed for our global optimization for the deformations and camera pose detailed in the next section.

\subsection{Tracking camera pose and deformation}

Our goal is, given some feature matches in the current frame $\mathbf{u}_i^t$ and the 3D reconstruction in the previous temporal instant $\mathbf{X}_i^{t-1}$, to estimate the current camera pose $\mathbf{T}_{C^tW}$ and the deformation $\pmb\delta_i^t$ of each point such as the current scene can be estimated as $\mathbf{X}_i^{t}=\mathbf{X}_i^{t-1}+{\pmb\delta}_i^t$.

For that we introduce a reprojection data term $E^t_{i,rep}$ along with 2 regularizers $E^t_{i,spa}$ and $E^t_{i,tmp}$ to constrain the deformations in our total global cost function, $\mathcal{E}^t$ for time $t$, defined by:

\begin{equation}
    \mathcal{E}^t = \sum_{i \in \mathcal{P}} E^t_{i,rep} + \lambda_{spa} E^t_{i,spa} + \lambda_{tmp} E^t_{t,tmp}
\end{equation}

where $\mathcal{P}$ represents the set of points being observed in the current frame. Our global problem can be solved using Non-Linear Squares optimization such us:

\begin{equation}
    \mathbf{T}_{C^tW}, 
    {\pmb\delta}_i^t = \underset{ \mathbf{T}_{C^tW}, 
    {\pmb\delta}_i^t}{\argmin} \ {\mathcal{E}^t}
\end{equation}

Next we define the terms of our cost function $\mathcal{E}^t$.

\subsubsection{Reprojection term}
we obtain feature matches $\mathbf{u_i^t}$ in the current frame with the modified Lucas-Kanade algorithm presented in \cite{gomez2021sd}, computing on this way the reprojection error as it follows:

\begin{equation}
E^t_{i,rep} = \rho \big(\| \mathbf{u}_i^t-\mathbf{\hat{u}}_i^t \|^2_{\Sigma{rep}} \big)
\end{equation}

where $\rho$ is the Hubber robust cost, $\mathbf{\hat{u}}_i^t$ and $\mathbf{u}_i^t$ are respectively the match of feature $i$ in the current image $I_t$ and  its projection given by:

\begin{equation}
\mathbf{\hat{u}}_i^t = \Pi(\mathbf{T}_{C^tC^0} (\Pi^{-1}(\mathbf{u}_i^0,d_{i}) + {\mathbf{X}_i^{t-1}+\pmb\delta}_i^t)
\end{equation}

The accuracy of indirect methods is limited by the feature detector resolution (typically no less than 1 pixel). However matches obtained with photometric methods have subpixel accuracy boosting on this way the accuracy of our reprojection term while keeping its nice convergence basin.

\subsubsection{Spatial regularizer}
Following \cite{newcombe2015dynamicfusion} we encode assumption (2) with a regularizer that constrains deformations locally so they are spatially smooth:

\begin{equation}
E^t_{i,spa} = \sum_{j \in \mathcal{G}(i)} \rho\big( \| w^t_{ij} ({\pmb\delta}_i^t  - {\pmb\delta}_j^t) \|^2_{\Sigma_{spa}} \big)
\end{equation}

Here $\mathcal{G}$ represents a weighted graph that encodes related points whose deformations should be regularized together. The weight in $\mathcal{G}$ of two connected points $i$ and $j$ is $w^t_{ij}$ which depends on the Euclidean distance between both points at the immediately previous time instant $t-1$ and is computed according to the following formula:

\begin{equation}
    w^t_{ij} = \exp\left({\frac{-\left\|\mathbf{X}_i^{t-1} - \mathbf{X}_j^{t-1} \right\|^2}{2 \sigma^2}}\right)
\end{equation}

where $\sigma$ is a radial basis weight that controls the influence radius of each point. This regularizer is crucial as it enforces as rigid-as-possible deformations and contributes towards a global consistency of the deformations.

\subsubsection{Temporal regularizer}
Finally we add a temporal regularizer on the deformations to represent that they occur slowly over time (assumption (3)):

\begin{equation}
E^t_{i,tmp} = \rho\big(\| {\pmb\delta}_i^t \|^2_{\Sigma_{tmp}} \big)
\end{equation}

This regularizer also interacts with assumption (4) as it penalizes big deformations that could be explained with a camera motion.

\section{EXPERIMENTS}\label{sec::experiments}
\begin{figure*}[]
\vspace{2mm}
\centering
    \begin{subfigure}{\linewidth}
        \centering
        \includegraphics[width=0.62\linewidth]{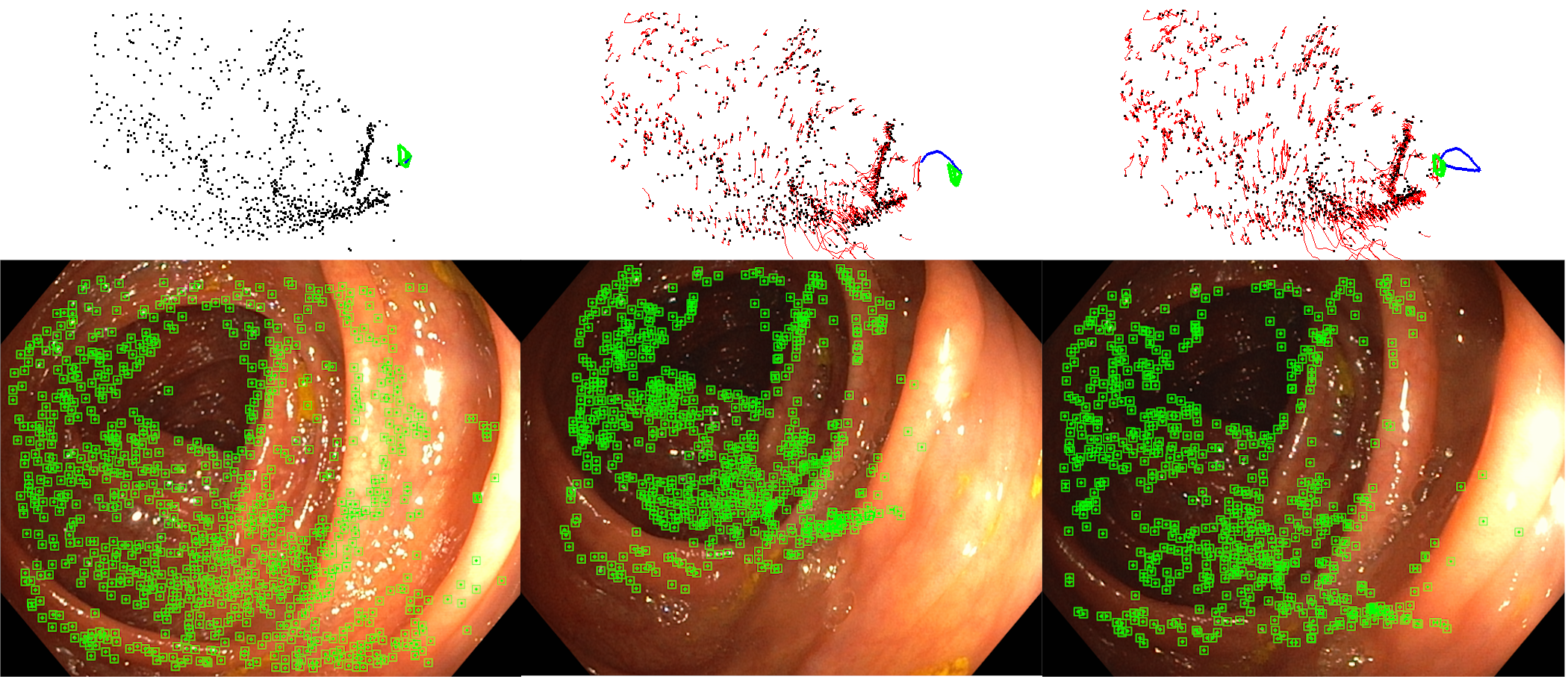}
        \caption{}
        \label{fig::48}
    \end{subfigure}\\
    \begin{subfigure}{\linewidth}
        \centering
        \includegraphics[width=0.62\linewidth]{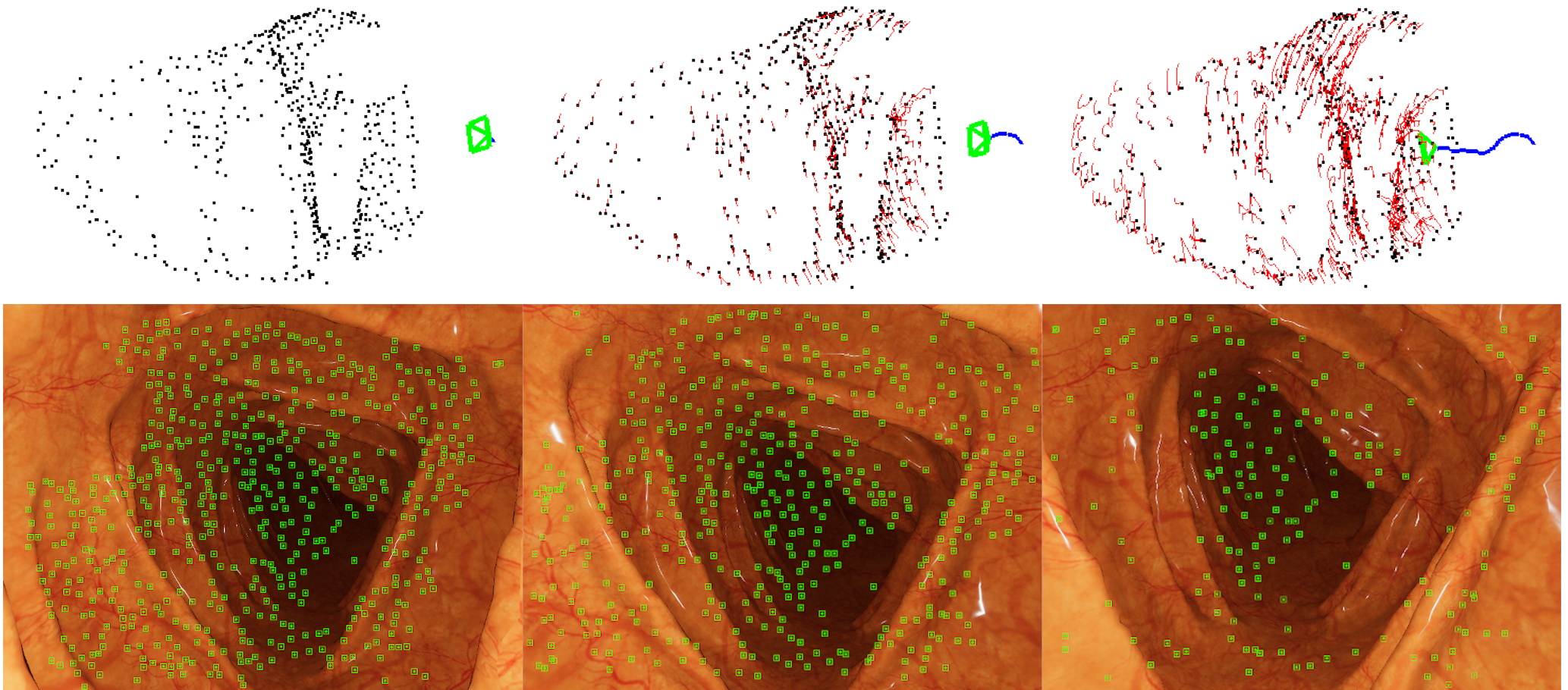}
        \caption{}
        \label{fig::simu}
    \end{subfigure}\\
    \begin{subfigure}{\linewidth}
        \centering
        \includegraphics[width=0.62\linewidth]{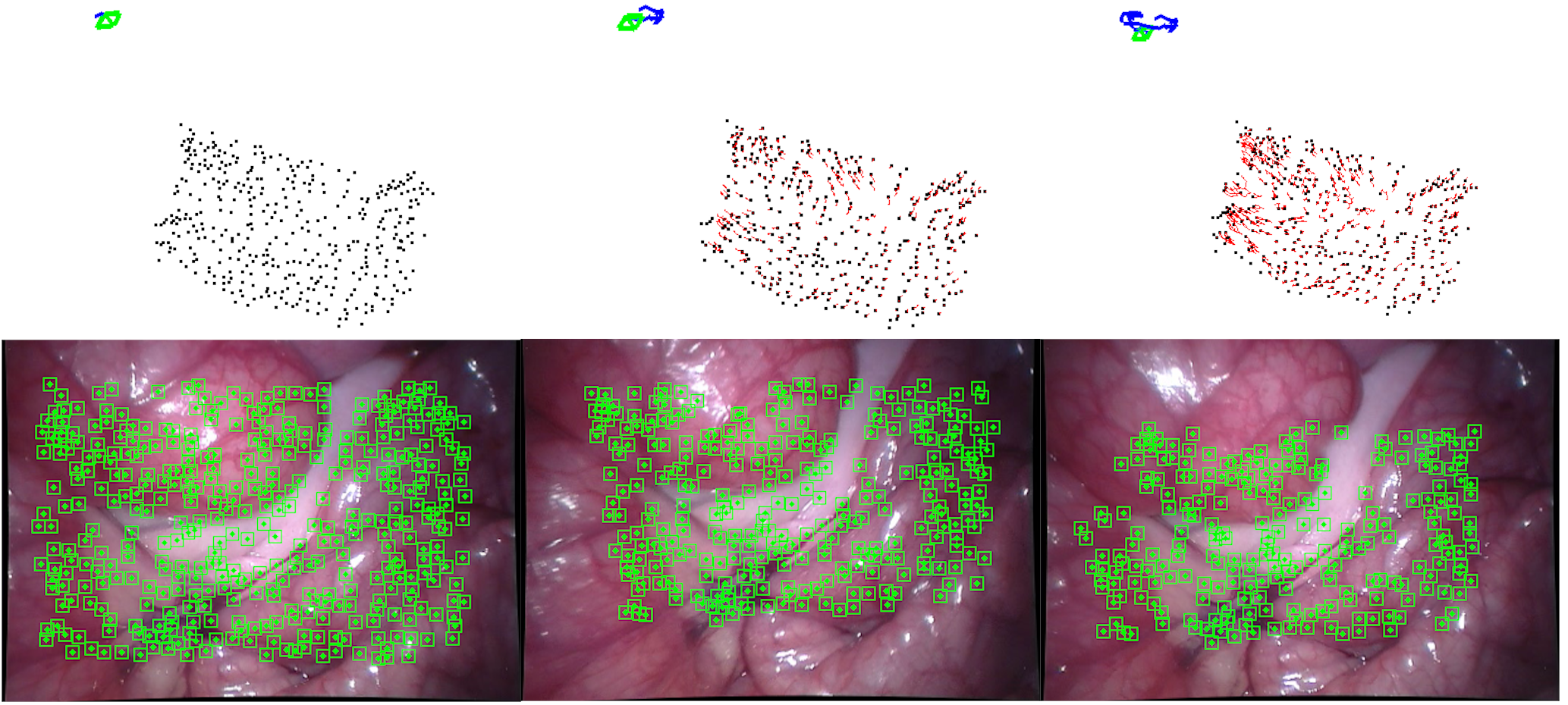}
        \caption{}
        \label{fig::20}
    \end{subfigure}\\
    \begin{subfigure}{\linewidth}
        \centering
        \includegraphics[width=0.62\linewidth]{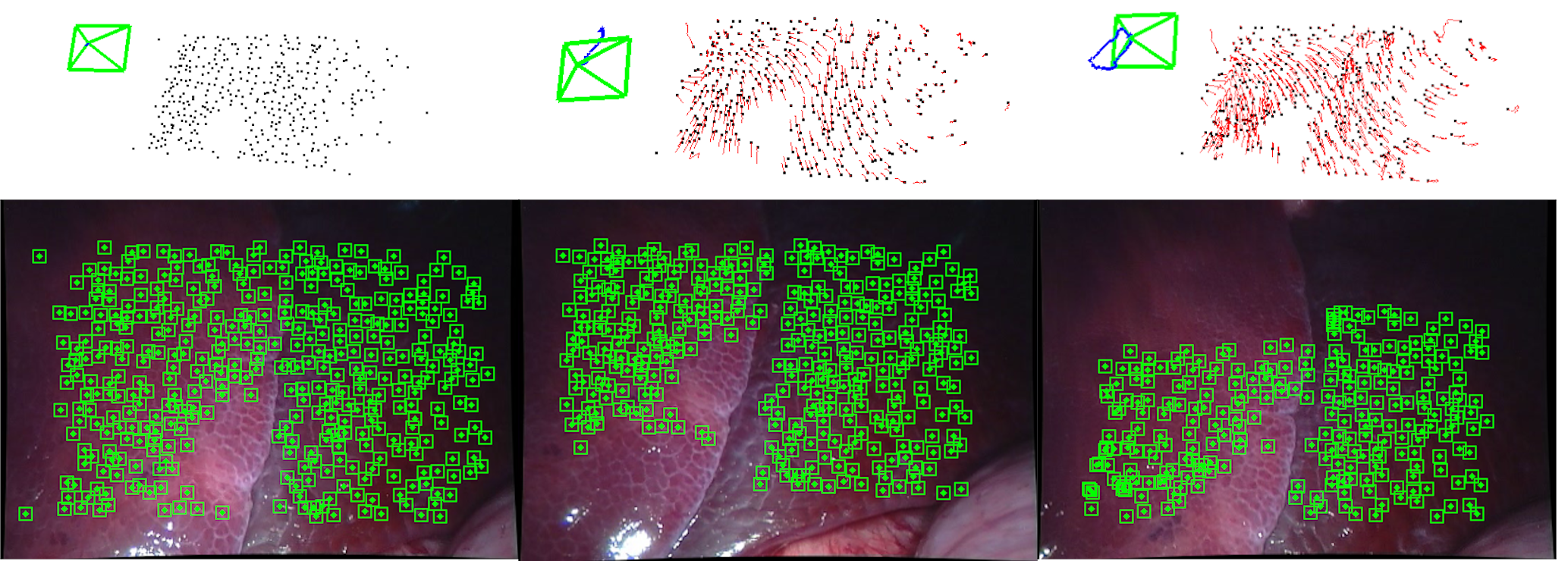}
        \caption{}
        \label{fig::21}
    \end{subfigure}\\

    \caption{Results of our algorithm for different sequences. Per each sequence, in columns results after the initial middle and final frame.  Two rows per sequence. The first row displays the 3D reconstruction: black points are the undeformed map, red lines are the map point deformation trajectories, in blue the camera trajectory. The second row the RGB frames with the tracked features in green. From top to bottom: (a) EndoMapper real in-vivo sequence, (b) Simulated sequence, (c) Hamlyn 20 sequence and (d) Hamlyn 21 sequence. All datasets have been processed using only monocular images.}\label{fig::all_datasets}
\end{figure*}

We evaluate our method in a Minimal Invasive Surgery sequences, more specifically in colonoscopies. This kind of sequences pose big challenges as they exhibit continuous deformations, poor textures and harsh illumination conditions. We provide quantitative results in photorealistic synthetic data and qualitative experiments with in-vivo human colonoscopies. For comparison purposes we also test our method in the Hamlyn dataset using its stereo setup to evaluate our reconstructions against other state-of-the-art methods. A summary of our main results can be seen in Fig \ref{fig::all_datasets}.



\subsection{Implementation details}
We implement the monocular map initialization, camera pose and deformation estimation in C++. For non Linear Squares optimization we use the Levenberg-Marquart algorithm implemented in the g2o library \cite{grisetti2011g2o}. For feature extraction and matching we implement our own Shi-Tomasi feature extractor and Lucas-Kanade tracker (Eq. \ref{eq::klt_enhan_residual}). We set a threshold of 0.8 for the SSIM score (Eq. \ref{eq::SSIM}) to detect and reject spurious feature tracks. Regarding the optimization, we set $\Sigma_{rep}$ to 1 pixel, $\Sigma_{spa}$ and $\Sigma_{tmp}$ to 10 mm. Since $\Sigma_{spa}$ and $\Sigma_{tmp}$ correctly scales the $E_{spa}$ and $E_{tmp}$ terms, we set their respective $\lambda$ to 1. Finally, for the Hubber cost threshold we use the 95 percentile of $\chi^2$, with 2 DoF for $E_{rep}$ and with 3 DoF for $E_{spa}$ and $E_{tmp}$.

Regarding the regularization graph $\mathcal{G}$, for each point we only add regularization terms with its $K=20$ closest points in 3D with $\sigma$ set to 15 mm when initializing the map with the monocular camera and 55 mm when using the stereo to get the first map reconstruction. This is done to ensure that a points is always regularized and at the same time ignoring points that have little influence with the current point to reduce the computational burden.

\subsection{Simulated Colon dataset}
We use the VR-Caps \cite{incetan2021vr} to generate photerealistic synthetic image sequences of a 3D colon model obtained from a Computed Tomography. Since this is a simulation, we have access full to camera pose and 3D scene ground truth. Indeed, we can generate sequences with different camera trajectories and degrees of deformation enabling us to test each one of the components of our system individually.

For evaluation purposes, we simulate an insertion maneuver (Fig. \ref{fig::simu}) with different degrees of deformation. We model the deformations via a sine wave propagating along the simulated colon. We apply this deformations to the $y$ coordinate of the point surfaces simulating perilstatic movements according to the following formula:

\begin{equation}
    V_y^t = V_y^0 + A \sin(\omega t + V_x^0 + V_y^0 + V_z^0)
\end{equation}

where $V_x^0$, $V_y^0$ and $V_z^0$ are the coordinates of the surface point at rest. We can control the magnitude and velocity of the deformations according to the parameters $A$ and $\omega$ respectively. Table \ref{tb::simulation} shows the reconstruction accuracy of our system in the simulated sequence with different deformation velocities and amplitudes. The error shown is the Root Mean Square Error (RMSE) of the reconstructed points for all frames according to:

\begin{equation}
e_{\text{rms}} = \sqrt\frac{\sum_i\|s^t\mathbf{\hat{X}}_i^t-\mathbf{X}^{t,{gt}}_i\|^2}{n}
\label{eq:rmse:perframe:alignment}
\end{equation}

Since this is a full monocular formulation, we find, for each frame, an optimal scale factor, $s^t$, to align our reconstructions with the ground truth.

\begin{table}[t]
\vspace{2mm}
\centering
\caption{Reconstruction RMSE (mm) in simulated colonoscopies \cite{ozyoruk2020endoslam} for different deformation types}\label{tb::simulation}
\begin{tabular}{|l||c|c|c|}
\hline
\backslashbox{$A$ (mm)}{$\omega$ (rad/s)}      & 0    & 2.5  & 5       \\ \hline\hline
0     & 1.15 & -    & -         \\ \hline
2.5 & -    & 1.77 & 1.70   \\ \hline
5  & -    & 1.84 & 3.65  \\ \hline
10   & -    & 2.27 & 4.57  \\ \hline
\end{tabular}
\end{table}

Results show that our formulation can reach nice reconstruction error around 2-3 mm even though in presence of deformations. One interesting result is that our system is more sensitive to deformation velocities than the magnitude itself being aligned with assumption (3).

\begin{table*}[t]
\vspace{2mm}
\centering
\caption{Comparison with previous methods in sequences 20 and 21 from Hamlyn Dataset as shown in \cite{lamarca2021direct}. \\We report reconstruction RMSE (mm) and number of frames processed.}\label{tb::hamlyn}
\begin{tabular}{cc|c|cccc|}
\cline{3-7}
                                          &        & \multicolumn{4}{c|}{Stereo Initialization} & Monocular Init.                                                                                                                                                                      \\ \cline{3-7} 
                                          &        & ORB-SLAM3 \cite{campos2021orb}  & \multicolumn{1}{c|}{SD-DefSLAM \cite{gomez2021sd}}  & \multicolumn{1}{c|}{DSDT \cite{lamarca2021direct}} & \multicolumn{1}{c|}{\begin{tabular}[c]{@{}c@{}}Ours\end{tabular}} & \begin{tabular}[c]{@{}c@{}}Ours\end{tabular} \\ \hline
\multicolumn{1}{|c|}{\multirow{2}{*}{20}} & RMSE   & 1.37      & \multicolumn{1}{c|}{4.68} & \multicolumn{1}{c|}{2.9}  & \multicolumn{1}{c|}{1.48}                                                        & 2.79                                                           \\ \cline{2-7} 
\multicolumn{1}{|c|}{}                    & \# Fr. & 220       & \multicolumn{1}{c|}{252}  & \multicolumn{1}{c|}{500}  & \multicolumn{1}{c|}{350}                                                         & 350                                                            \\ \hline\hline
\multicolumn{1}{|c|}{\multirow{2}{*}{21}} & RMSE   & -         & \multicolumn{1}{c|}{6.19} & \multicolumn{1}{c|}{1.3}  & \multicolumn{1}{c|}{1.55}                                                        & 3.31                                                           \\ \cline{2-7} 
\multicolumn{1}{|c|}{}                    & \# Fr. & -         & \multicolumn{1}{c|}{323}  & \multicolumn{1}{c|}{300}  & \multicolumn{1}{c|}{300}                                                         & 300                                                            \\ \hline
\end{tabular}
\end{table*}

\subsection{Hamlyn dataset}
We also test our formulation in real endoscopic sequences. For that purpose, we use sequences 20 (Fig. \ref{fig::20}) and 21 (Fig. \ref{fig::21}) of the Hamlyn dataset \cite{mountney2010three}. Sequence 20 (from frame \#750) corresponds to  abdominal exploration with slow deformation. Sequence 21 (also from frame \#750) images a liver with 2 lobes each of them moving on its own. This can be considered as an articualted motion. In both sequences, surface texture is poor and illumination conditions are unfavorable. This dataset is recorded with a stereo endoscope, allowing us to estimate environment groundtruth from the disparity observed by the stereo sensor.

We evaluate our formulation in 2 setups (Table \ref{tb::hamlyn}) for comparison purposes. In the first setup, we initialize our system with the first stereo images and perform monocular tracking, in order to allow comparison with previous methods ORB-SLAM \cite{mur2015orb}, SD-DefSLAM \cite{gomez2021sd} and  Direct and Sparse Deformable Tracking (DSDT) \cite{lamarca2021direct}. Since we are initializing from the stereo images, we do not perform any scale alignment when computing the RMSE. We achieve competitive results regarding reconstruction error, obtaining a consistent error around 1.5 mm. Since we do not impose any restriction on the surface topology or in the deformations, we achieve a significantly smaller error compared with SD-DefSLAM. This is specially clear in sequence 21 with the 2 lobes moving independently what limits the accuracy of SD-DefSLAM.
The comparison with DSDT suggests that our regularizers are versatile, we are able to code better the spatial smoothness of sequence 20,  achieving a lower error, while still being competitive in hard discontinuity of sequence 21. DSDT is able to keep the track longer because, in contrast with DSDT, our method still does not implement any policy to recover points lost during tracking.

The second setup uses the full monocular pipeline including our monocular initalization (Sec.\,\ref{sec:monocular:initialization}) computing the RMSE after a per frame scale correction (Eq.\,\ref{eq:rmse:perframe:alignment}). In this scenario, our system reaches errors around 2.8-3.3 mm which is aligned with the errors obtained in the simulation dataset under significant deformations. The increase in error compared with the stereo setup is due to the quality of the map initialization that no longer relies on a perfect stereo initialization.

Also it is important to note that in these sequences the surfaces shape and deformations observed are completely different from the ones seen in the simulation dataset proving that we can model general surface shapes and deformations.

\subsection{Real endoscopy sequences}
We provide qualitative results in real in-vivo human colonoscopy sequences from the EndoMapper project \cite{Endomapper}. These sequences display the big challenges real colonoscopies pose, such as deformation, little to no texture in the images, lighting conditions varying from frame to frame, reflections and fish-eye optics (fig.\, \ref{fig:Tracking_Endomapper}.) 

In this case, there is no ground truth to compare with, because the dataset just records standard monocular endoscope procedures. For this reason, we only provide qualitative results. Fig.\, \ref{fig::init} displays how we are able to initialize maps with high density of points form quite close (3 frames apart) frames capturing the tubular topology of the colon. In Figures \ref{fig:Tracking_Endomapper} and \ref{fig::48}, it can be seem how our algorithm is able to capture the scene deformation and the endoscope trajectory, being able to track map points for more than 30 frames in two examples of real colonoscopies of two different patients.





\section{CONCLUSIONS}\label{sec::conclusions}

In this work we have presented an approach for monocular camera tracking and deformation estimation without assumptions on the environment shape or topology.  Instead, we successfully encode with simple regularizers the assumptions about the type of deformations that are common in endoscopy. Compared with the state of the art, and  our method, including map initialization, is applicable in a much wider range of shape and topologies, like colonoscopies, while having similar accuracy in more standard almost-planar scenarios. 


The presented monocular initialization and tracking contributes to make real a fully deformable SLAM system. The deformable mapping to expand the map as the camera explores new regions is closer after our contribution, being a promising venue of future work in the short term. In the mid-term, a multi-map deformable SLAM offers a profitable for future work  because will be able to cope with occlusions and tracking losses prevalent in real colonoscopies.

\addtolength{\textheight}{-12cm}   


\bibliographystyle{IEEEtran}
\bibliography{IEEEabrv,bibl}

\end{document}